\documentclass[runningheads]{llncs}
\usepackage[T1]{fontenc}
\usepackage{hyperref}
\usepackage{subcaption}
\usepackage{graphicx,verbatim}
\usepackage{amsmath}
\usepackage{amsfonts}
\usepackage{caption}
\begin{document}
\title{Orientation-Aware Unsupervised Domain Adaptation for Brain Tumor Classification Across Multi-Modal MRI}
%
\author{
Sapna Sachan\inst{1}  \and
Amulya Kumar Mahto\inst{1}
Prashant Wagambar Patil \inst{1}
}

\authorrunning{F.~Author et al.}

\institute{Indian Institute of Technology, Guwahati \email{akmahto@iitg.ac.in}\\}

\maketitle              
\begin{abstract}
The clinical integration of deep learning models for brain tumor diagnosis in neuro-oncology is severely constrained by limited expert-annotated MRI data and substantial inter-institutional domain shift arising from variations in scanners, imaging protocols, and contrast settings. These challenges significantly impair model generalization in real-world settings. To address this, we propose a novel orientation-aware unsupervised domain-adaptive framework for automated brain tumor classification using mixed 2D MRI slices. Initially, a CNN with large receptive field first categorizes input slices into axial, sagittal, and coronal views. For each orientation, a CNN architecture with ResNet50 backbone augmented with four fully connected layers is trained to extract discriminative features for tumor classification. To mitigate annotation scarcity and domain discrepancies, we introduce a slice-wise unsupervised domain adaptation strategy that transfers knowledge from the multi-modal (T1, T2, and FLAIR) source domain to the single-modality (post-contrast T1) target domain. Feature-level alignment is enforced using maximum mean discrepancy loss, complemented by pseudo-label–guided adaptation to preserve class discriminability. Extensive experiments demonstrate improved target-domain performance over prior approaches, highlighting the benefits of orientation-specific learning, multi-modal knowledge transfer, pseudo-label-guided adaptation, and unsupervised domain adaptation.

\end{abstract}
\section{Introduction}
MRI is the primary modality for brain tumor diagnosis due to its non-invasive nature and superior soft-tissue contrast, with multi-sequence acquisitions (T1, T2, FLAIR) and multi-planar views (axial, sagittal, coronal) providing complementary anatomical and pathological information~\cite{westbrook2018}. Despite advances in deep learning algorithms for brain tumor classification~\cite{ali2025multi}, clinical adoption remains limited by scarce expert annotations and domain shifts arising from variations in scanners, acquisition protocols, and institutions~\cite{kushol2023dsmri}. These factors cause models trained on labeled source data (from one institution/scanner) to generalize poorly to unlabeled target data (from a different institution/scanner)~\cite{guo2024scanner_shift}. Unsupervised domain adaptation (UDA) offers a promising solution for the domain shift problem. By learning domain-invariant yet class-discriminative representations, UDA transfers knowledge from a labeled source domain to an unlabeled target domain \cite{guan2022survey}. However, existing UDA methods struggle for multi-class brain tumor classification with complex inter-sequence and multi-orientation variability as they are predominantly designed for binary classification and non-complex domain shifts and
To address these challenges, we propose a novel orientation-aware UDA framework for multi-class brain tumor classification, explicitly addressing annotation scarcity, domain shift, and orientation-specific discrepancies. Our key contributions are:
\begin{enumerate}
\item We design a CNN-based slice separation module with large receptive field that classifies MRI slices into axial, sagittal, and coronal orientations, enabling orientation-specific learning.
\item The work employs orientation-specific classifiers with ResNet50 backbone which learns class-discriminative features in each orientation independently.
\item The paper introduces a pseudo-label–guided, class-wise domain adaptation strategy that aligns source and target class features using MMD with a Gaussian kernel, preserving class-specific semantics while mitigating domain shift.
\end{enumerate}

\section{Related Work}

Deep learning has substantially advanced automated brain tumor analysis, including segmentation, detection, and classification~\cite{ali2025multi}. Early approaches relied on handcrafted features, whereas modern methods predominantly use CNNs and transfer learning to extract hierarchical representations from MRI data~\cite{swati2019,PRABHAS2025200389,app152413155}. Multi-class tumor classification has benefited from powerful backbones such as VGG, ResNet, and EfficientNet, as well as attention-based and hybrid ensemble models~\cite{sharma2023brain,shamshad2024enhancing,islam2023fine,apostolopoulos2023attention,mozumdar2025efficient}.Despite strong performance on curated datasets, these supervised methods exhibit limited generalization under domain shift, particularly in multi-center and multi-plane settings.
UDA mitigates this limitation by learning domain-invariant representations from source and target data and needs only source annotations ~\cite{guan2022survey}. Classical approaches include discrepancy-based methods such as Deep Domain Confusion (DDC)~\cite{tzeng2014deep} and Joint Adaptation Networks (JAN)~\cite{long2017jan}, which align source and target features in latent space using Maximum Mean Discrepancy. Adversarial methods, including DANN~\cite{ganin2016dann} and CDAN~\cite{long2018cdan}, promote domain-invariance through gradient reversal and conditional alignment. More recent strategies incorporate class-wise or semantic alignment via decision-boundary disagreement (MCD)~\cite{saito2018mcd}, domain-specific normalization (DSBN)~\cite{chang2019dsbn}, and source-free adaptation frameworks such as SHOT~\cite{liang2020shot} and DJSA~\cite{djsa}.
In medical imaging, UDA has been primarily explored for segmentation tasks, as highlighted by crossMoDA challenges and recent surveys~\cite{crossMoDA2022,uda_seg_review}. Classification-focused UDA studies remain limited. Tang et al.~\cite{tuna_net} proposed an adversarial UDA framework for pneumonia detection in chest X-rays, restricted to binary classification of single-view images. Pseudo-label–based UDA has been applied to COVID-19 detection for cross-domain generalization, though restricted to binary classification~\cite{covid_pseudo_uda}. In neuroimaging, Li et al.~\cite{cross_modal_brain_da} investigated cross-modal adaptation between CT and MRI for tumor versus non-tumor classification.

\textbf{Research Gap:} UDA for multi-class brain tumor classification is unexplored due to cross-institutional MRI variability and scarce annotations. Existing methods ignore class-wise semantics and orientation-specific shifts. We address this gap via an orientation-aware, class-wise adaptation framework for robust multi-class MRI tumor classification.

\section{Description of Dataset}
\begin{figure}[t]
    \centering
    \begin{minipage}[t]{0.48\textwidth}
        \centering       \includegraphics[width=\linewidth]{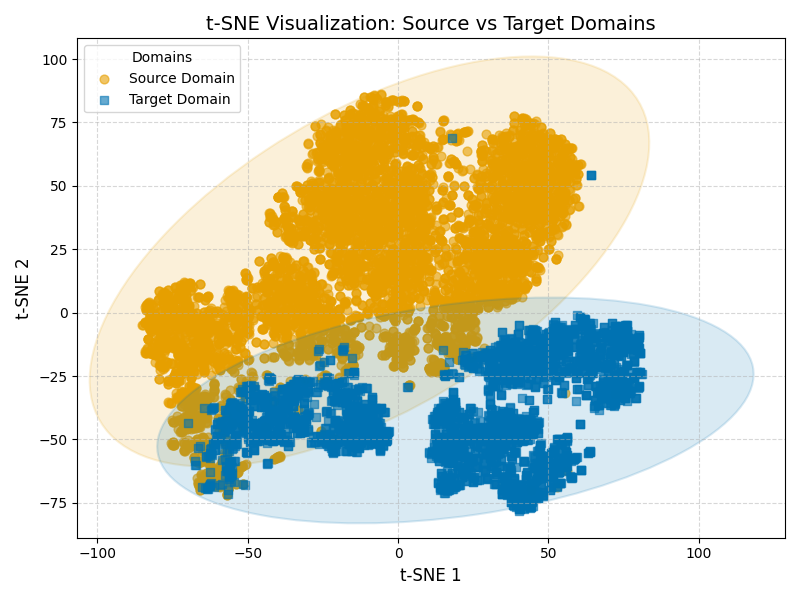}
        \caption{t-SNE visualization showing feature distribution shift between source and target domains.}
        \label{fig:tsne}
    \end{minipage}
    \hfill
\begin{minipage}[t]{0.49\textwidth}
        \centering
        \includegraphics[width=\linewidth]{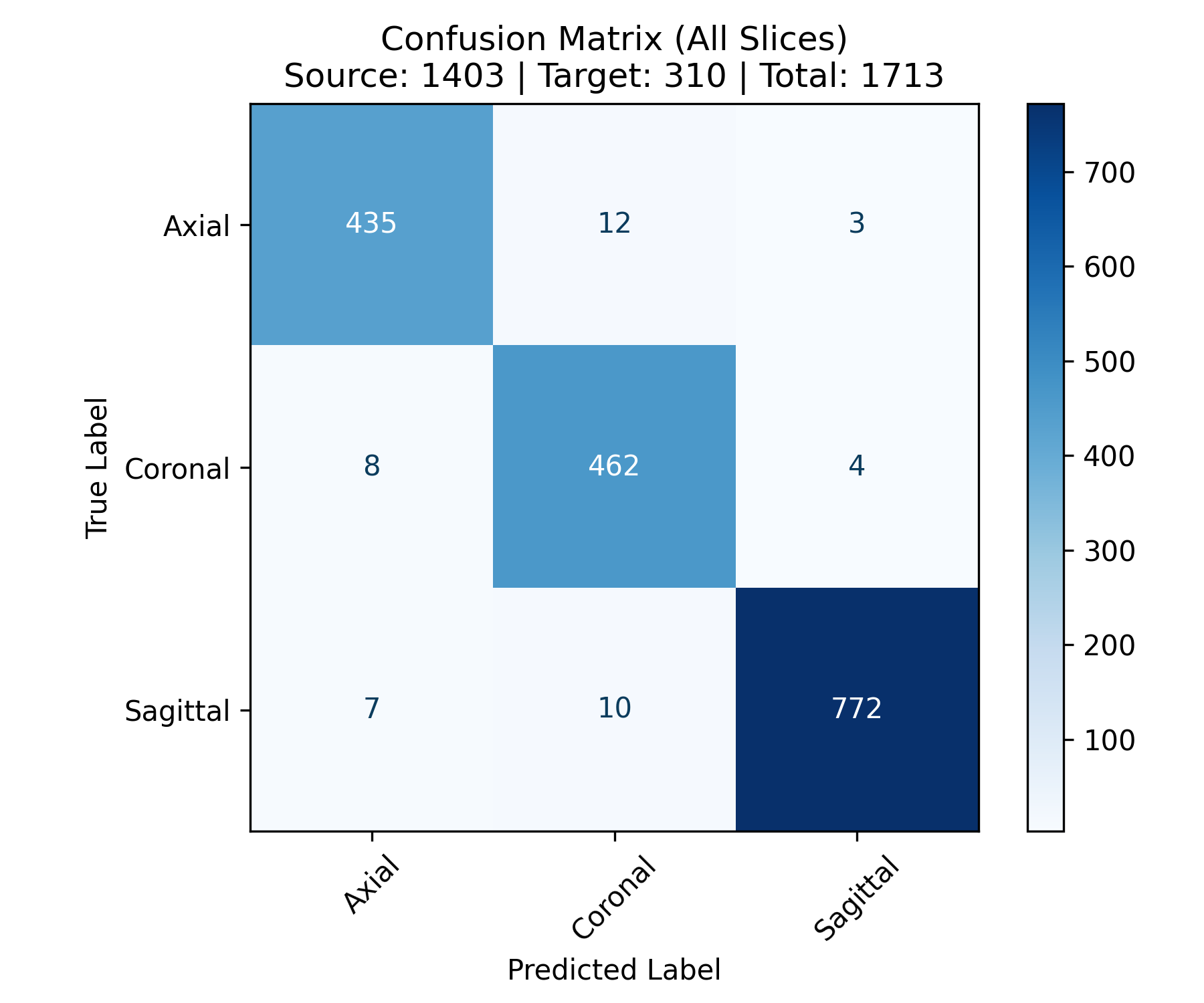}
        \caption{Confusion matrix of the slice separation model evaluated on combined Source and target test slices}
\label{fig:confusion}
    \end{minipage}
\end{figure}

We use two datasets in this study: source dataset and target dataset. The source dataset is the \textit{Bangladesh Brain Cancer MRI Dataset}~\cite{bangladesh_brain_mri}, a multi-modal collection comprising T1, T2, and FLAIR-weighted MRI scans that capture complementary tissue characteristics. It contains 6,056 2D MR images across three tumor classes: glioma, meningioma, and pituitary. The target dataset is the \textit{Figshare Brain Tumor Dataset} ~\cite{cheng_brain_tumor}, consisting of 3,064 post-contrast T1-weighted MR images of 223 patients. The target dataset includes 2D slices extracted from 3D volumes and shares the same three tumor categories as the source dataset. Both datasets contain mixed anatomical planes, and their class-wise slice distributions are as shown in Table~\ref{tab:class_slice_distribution}. A t-SNE visualization of extracted features (Fig.~\ref{fig:tsne}) demonstrates a pronounced domain gap between the two datasets, motivating the use of Domain Adaptation(DA) for effective feature alignment.

\section{Proposed Framework}
This section discribes a two-stage framework for multi-class brain tumor classification from MRI scans, including data preparation, pre-processing, model training, and evaluation. The framework consists of (i) DilatedCNN-based slice separation and (ii) orientation-specific tumor classification with UDA.

\subsection{Stage 1: Slice Separation Model}
To provide a structured representations of MR images having mixed orientations, we first leverage a \textit{slice separation model} that classifies MRI slices into their anatomical orientations: \textbf{axial}, \textbf{sagittal}, and \textbf{coronal}. This step reduces the overall complexity of the domain shift, which is inherently greater in data with mixed orientations than in any single orientation-specific subset therefore, providing a robust foundation for subsequent tumor classification.

\subsubsection*{Preprocessing and Binary Thresholding}
Given an image converted to grayscale,  \( I \in \mathbb{R}^{H \times W \times 1} \), we apply a \textit{binary thresholding} to suppress background noise. Mathematically, the transformation is defined as:
\begin{equation}
B(x,y) = 
\begin{cases} 
255, & I(x,y) > T, \\
0, & I(x,y) \leq T,
\end{cases}
\end{equation}
where \( T \) is the intensity threshold empirically set to \( T=35 \). The image is then resized to $32\times 32$ and normalized after thresholding. The resulting binary image  preserves only the boundary of the anatomical structures which is relevant for orientation classification while reducing non-informative background noise.

\subsubsection*{Dilated Convolutional Neural Network (DilatedCNN)}
The processed slices are then passed into a \textit{Dilated Convolutional Neural Network (DilatedCNN)} designed to extract both local and global contextual features. The model begins with standard convolutional blocks for local feature extraction, followed by dilated convolutions with dilation rates of 2 and 4 to enlarge the receptive field. This enables the network to capture local features while integrating global context. 

For this stage, a subset of both target and source were \textbf{manually annotated} as axial, coronal and sagittal orientations to provide ground truth labels. The classifier was trained on 350 source images and tested on a combined set of 1,713 source and target slices, enabling orientation-specific features that generalize across domains. As shown in Fig.~\ref{fig:model}, input MRI slices pass through local and global feature extractors to a classification head for accurate orientation separation. The network is trained using \textit{cross-entropy loss} for 50 epochs with SGD (learning rate 0.001, momentum 0.9).

\subsection{Stage 2: Classification Model with Domain Adaptation}
In the second stage, we address the problem of domain shift between the source and target datasets by adopting a \textit{two-phase unsupervised domain adaptation (UDA) strategy}. The source domain provides labeled images for supervised learning, whereas the target domain remains unlabeled and is incorporated through pseudo-labeling. To account for variations in anatomical orientations, the slices separated in Stage 1 are used to train three independent classification networks corresponding to \textbf{axial}, \textbf{sagittal}, and \textbf{coronal} views.

\subsubsection*{Model Architecture}
The classification model is built on a ResNet50 \cite{he2016deep} backbone pre-trained on ImageNet, where the convolutional feature extractor is frozen and followed by four fully connected layers of sizes $2048 \to 1024 \to 512 \to 256 \to 3$. ReLU activations are used in the hidden layers. The output layer generates predictions for the three tumor classes: \textit{glioma}, \textit{meningioma}, and \textit{pituitary}.

\subsubsection*{Phase 1: Supervised Training on Source}
In the first phase, the network is trained solely on the labeled source dataset using the categorical cross-entropy loss:
\begin{equation}
\mathcal{L}_{cls}^{(s)} = -\sum_{i=1}^{n_s} \sum_{c=1}^{C} y_{i,c}^{(s)} \log \hat{y}_{i,c}^{(s)} ,
\end{equation}
where, $y_{i,c}^{(s)}$ and $\hat{y}_{i,c}^{(s)}$ represent the ground truth and predicted probabilities for the source sample $i$ and class $c$. The trained source model is then used to assign \textbf{pseudo-labels} to the unlabeled target dataset, thereby enabling class-wise adaptation in the next phase.  

\begin{figure}[htbp]
    \centering
    \includegraphics[width=0.7\textwidth]{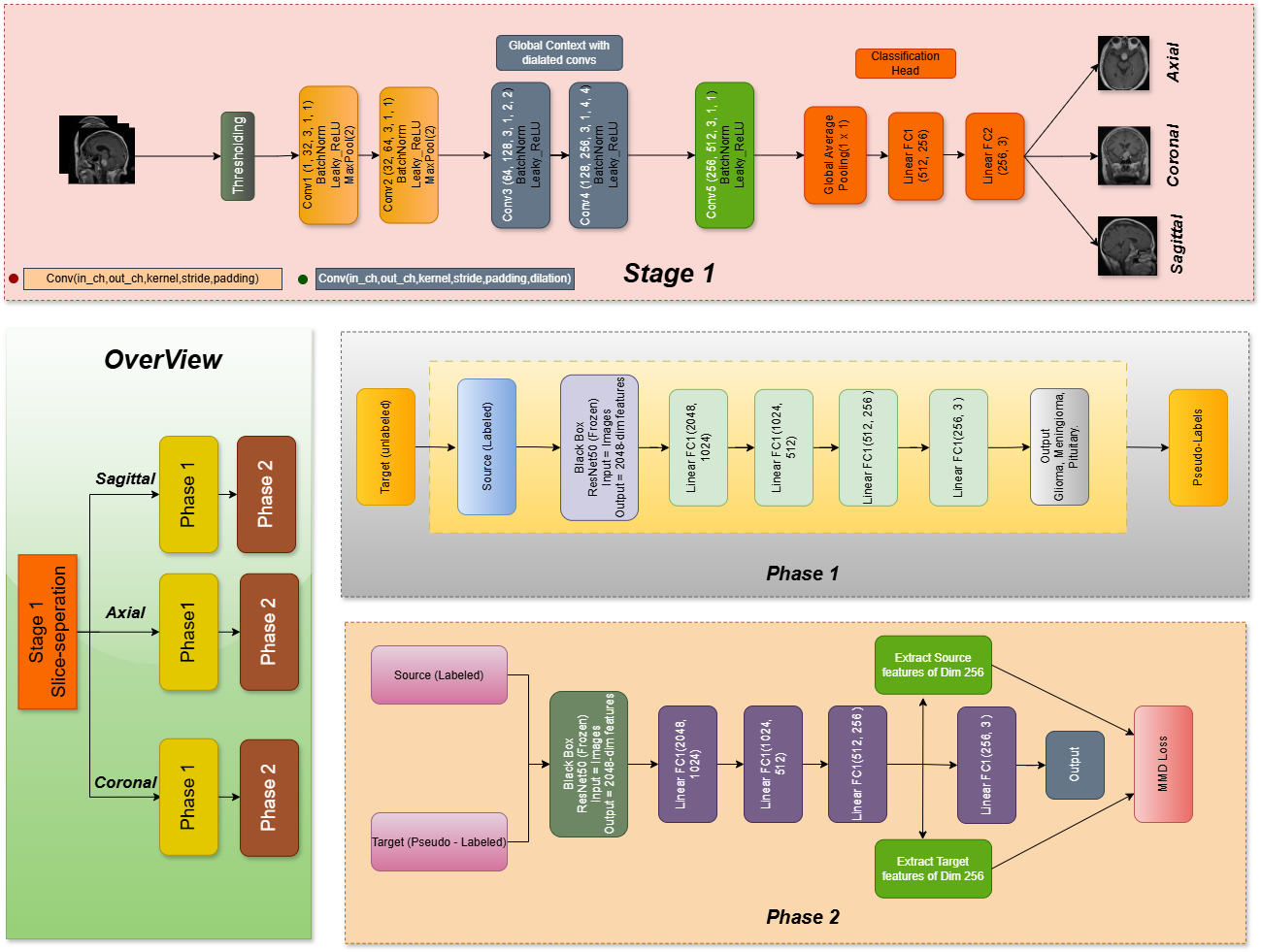}
    \caption{\textbf{Proposed Orientation-Aware UDA Framework:}
Stage 1 first classifies MRI slices into orientations using a DilatedCNN. Subsequently, Phase 1 trains orientation-specific ResNet50 models on labeled source data to generate pseudo-labels for the target domain, followed by Phase 2, which performs feature-level alignment between source and target representations to improve cross-domain tumor classification.}
    \label{fig:model}
\end{figure}

The architecture employed in Phase~1 is illustrated in Fig.~\ref{fig:model}, where the supervised model trained on the source domain generates pseudo-labels for the target domain, forming the basis for subsequent adaptation.

\subsubsection*{Phase 2: Class-wise Unsupervised Domain Adaptation}
In the second phase, we refine the model using both source and pseudo-labeled target data. The training objective combines classification losses from source and target domains with a feature alignment loss based on Maximum Mean Discrepancy (MMD):
\begin{equation}
\mathcal{L}_{total} = \mathcal{L}_{cls}^{(s)} + \mathcal{L}_{cls}^{(t)} + \lambda \, \mathcal{L}_{MMD},
\end{equation}
where,
$$
\mathcal{L}_{cls}^{(t)} = -\sum_{j=1}^{n_t} \sum_{c=1}^{C} \tilde{y}_{j,c}^{(t)} \log \hat{y}_{j,c}^{(t)} ,
$$
uses pseudo-labels $\tilde{y}_{j,c}^{(t)}$ for the target samples. Here, $\lambda$ is a trade-off hyperparameter that balances the importance of feature alignment (MMD loss) relative to the classification objectives. 

The MMD loss enforces alignment of feature distributions between the two domains in a reproducing kernel Hilbert space (RKHS) using a Gaussian kernel:
\begin{equation}
\mathcal{L}_{MMD} = \left\| \frac{1}{n_s}\sum_{i=1}^{n_s} \phi(x_i^s) - \frac{1}{n_t}\sum_{j=1}^{n_t} \phi(x_j^t) \right\|_{\mathcal{H}}^2 .
\end{equation}

An overview of the Phase~2 architecture is provided in Fig.~\ref{fig:model}, where class-wise domain adaptation is achieved by aligning source and pseudo-labeled target feature distributions through MMD loss.  

\subsubsection*{Training Setup}
After slice separation, class imbalance (Table \ref {tab:class_slice_distribution}) was mitigated using rotation-based augmentation in Sagittal slices of the source domain Glioma class. The source dataset was split 80:20 for training and validation, with target labels used only for testing. The model is trained for 20 epochs in Phase~1 and 100 epochs in Phase~2 using Adam (learning rate 0.001, momentum 0.9) with a fixed seed for reproducibility. The trade-off parameter $\lambda = 5 \times 10^{-4}$ was determined empirically to balance classification and alignment, applied independently to axial, sagittal, and coronal classifiers to form the final orientation-aware system.



\begin{figure*}[t]
    \centering

    \begin{subfigure}[t]{0.3\textwidth}
        \centering
        \includegraphics[width=\linewidth]{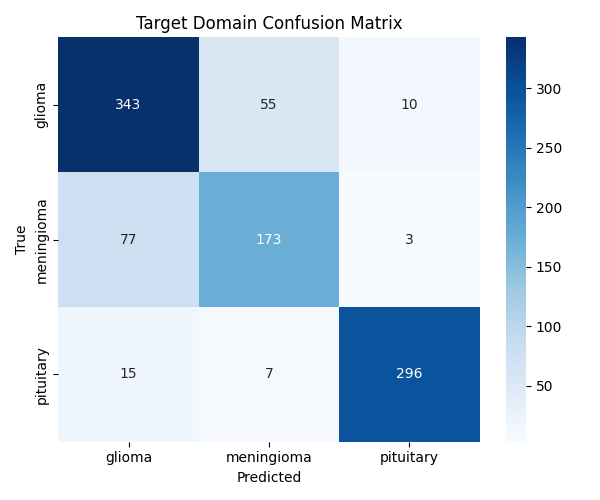}
        \caption{}
        \label{fig:sagittal}
    \end{subfigure}
    \hfill
    \begin{subfigure}[t]{0.3\textwidth}
        \centering
        \includegraphics[width=\linewidth]{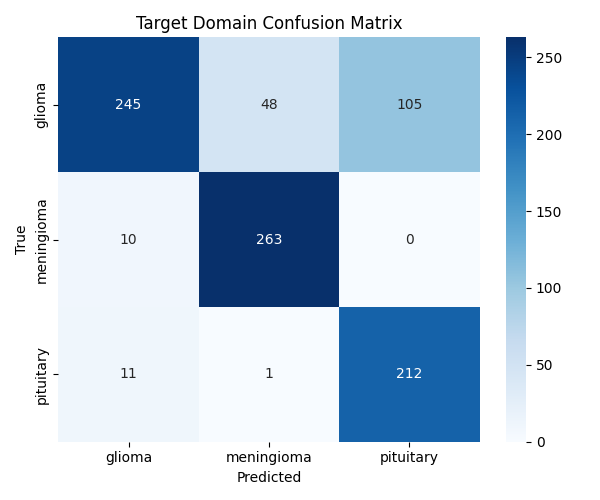}
        \caption{}
        \label{fig:coronal}
    \end{subfigure}
    \hfill
    \begin{subfigure}[t]{0.3\textwidth}
        \centering
        \includegraphics[width=\linewidth]{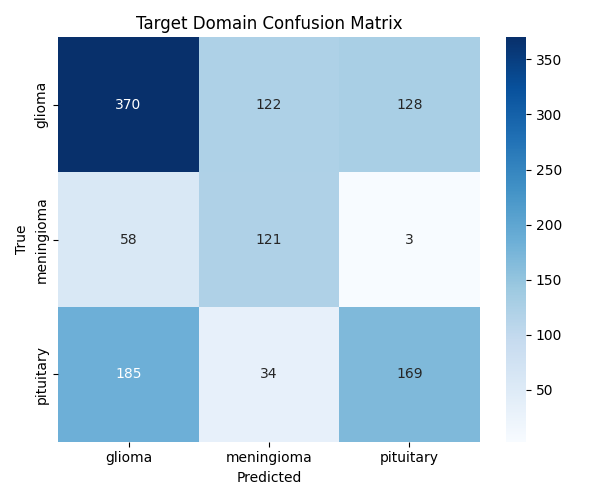}
        \caption{}
        \label{fig:axial}
    \end{subfigure}

    \caption{Target-domain confusion matrices using the proposed model: 
    (a) sagittal slices, 
    (b) coronal slices, and 
    (c) axial slices.}
    \label{fig:confusion_matrices}
\end{figure*}
\section{Results}
Performance is assessed using Accuracy and Macro F1-score. \textbf{Macro F1-score} averages class-wise F1 values without weighting by class frequency, providing a balanced measure of performance in multi-class and imbalanced settings.
\paragraph{\textbf{Stage 1: Slice Separation}}  
The Slice Separation model achieved a accuracy of $97.43\%$ on source and $97.42\%$ on target, demonstrating robust orientation classification. Table~\ref{tab:class_slice_distribution} details class-wise distributions and Confusion matrices (Fig.~\ref{fig:confusion}) on combined source and target test slices confirm accurate and balanced predictions across axial, sagittal, and coronal orientations.

\begin{table*}[t]
\centering
\begin{minipage}[t]{0.45\textwidth}
\centering
\caption{Macro F1-Score($\%$) 
performance without DA.}
\label{tab:baseline}
\begin{tabular}{|l|c|c|}
\hline
\textbf{Model} & \textbf{Source} & \textbf{Target} \\
\hline
EfficientNetB0 & 99.99 & 37.43 \\
DenseNet121    & 99.97 & 36.95 \\
ConvNeXt-Tiny  & 99.93 & 38.74 \\ 
ResNet50       & 99.92 & 37.96 \\
Transfusion    & 97.94 & 37.89 \\
MONAI ViT      & 95.29 & 37.95 \\
Medclip        & 89.87 & 37.96 \\
ViT-B16        & 79.16 & 38.38 \\
\hline
\end{tabular}
\end{minipage}
\hfill
\begin{minipage}[t]{0.48\textwidth}
\centering
\caption{Macro F1-Score($\%$) comparison of UDA methods.}
\label{tab:comparison}
\begin{tabular}{|l|c|c|}
\hline
\textbf{Approach} & \textbf{Source} & \textbf{Target} \\
\hline
SHOT~\cite{liang2020shot}                   & 96.78 & 38.48 \\
DDC~\cite{tzeng2014deep}                      & 96.20 & 38.64 \\
DJSA~\cite{djsa}                    & 89.03 & 47.39 \\
JAN ~\cite{long2017jan}                     & 68.15 & 49.90 \\
CDAN ~\cite{long2018cdan}                    & 85.17 & 50.46 \\
\textbf{W/O slice sep(Ours) }  & \textbf{98.43} & \textbf{51.96} \\
DANN ~\cite{ganin2016dann}                    & 52.44 & 54.28 \\
MCD  ~\cite{saito2018mcd}                    & 62.95 & 54.99 \\
DSBN ~\cite{chang2019dsbn}                    & 57.51 & 56.89 \\
\hline
\textbf{Ours} & \textbf{91.75} & \textbf{72.95} \\
\hline
\end{tabular}
\end{minipage}
\end{table*}

\paragraph{\textbf{Stage 2: Classification with and without DA}}  
Models trained without DA generalize poorly to the target domain, achieving only $\approx 38\%$ target Macro F1-score due to severe inter-domain discrepancies (Table~\ref{tab:baseline}). All models were initialized with publicly available pretrained weights, fine-tuned on the source dataset, and directly evaluated on the target dataset without adaptation. While existing UDA methods provide measurable improvements, their target performance remains limited ($\leq 57\%$), indicating insufficient alignment of orientation-specific and class-discriminative representations (Table~\ref{tab:comparison}). For fair comparison, all UDA methods, including our proposed framework, employ the same ResNet50 backbone for feature extraction; therefore, performance differences arise from the adaptation strategy rather than variations in network capacity.
Our approach substantially outperforms all competing UDA methods by explicitly combining orientation-aware slice separation with pseudo label-guided MMD alignment, achieving $73\%$ target Macro F1-score. Notably, accuarcy of the model on target domain degrades to $52\%$, without slice seperation confirming that orientation-specific modeling is a key component for the performance gains beyond conventional domain alignment. Confusion matrices Fig.~\ref{fig:sagittal} and  Fig.~\ref{fig:coronal}
further reveal balanced predictions for coronal and sagittal views, while axial model Fig.~\ref{fig:axial}
retain minor ambiguities, underscoring the necessity of orientation-aware learning. Overall, these results demonstrate that slice separation is critical component for learning domain-invariant yet class-discriminative features, effectively mitigating domain shift and class imbalance in multi-class brain tumor classification.

\begin{table*}[t]
\centering

\begin{minipage}[t]{0.33\textwidth}
\centering
\caption{Slice-wise Macro F1-Score($\%$) of the proposed method on Source and target domains.}
\label{tab:slice_accuracy}
\begin{tabular}{|l|c|c|}
\hline
\textbf{Slice} & \textbf{Source} & \textbf{Target } \\
\hline
Coronal   & 96.94 & 80.45 \\
Sagittal  & 89.79 & 82.94 \\
Axial     & 88.52 & 55.46 \\
\hline
\textbf{Avg} & \textbf{91.75} & \textbf{72.95} \\
\hline
\end{tabular}
\end{minipage}
\hfill
\begin{minipage}[t]{0.64\textwidth}
\centering
\caption{Class-wise and slice-wise distribution of Source and Target datasets after slice separation.}
\label{tab:class_slice_distribution}
\resizebox{\textwidth}{!}{
\begin{tabular}{|c|c|ccc|c|}
\hline
\textbf{Orientation} & \textbf{Domain} &
\multicolumn{3}{c|}{\textbf{Class-wise Distribution}} &
\textbf{Total} \\
\cline{3-5}
 &  & \textbf{Glioma} & \textbf{Meningioma} & \textbf{Pituitary} &  \\
\hline
Axial
 & Source & 1184 & 717 & 535 & \textbf{2436} \\
 & Target & 620 & 182 & 388 & \textbf{1190} \\
\hline
Sagittal
 & Source & 68 & 762 & 834 & \textbf{1664} \\
 & Target & 408 & 253 & 318 & \textbf{979} \\
\hline
Coronal
 & Source & 752 & 525 & 679 & \textbf{1956} \\
 & Target & 398 & 273 & 224 & \textbf{895} \\
\hline
\multicolumn{2}{|c|}{\textbf{Total}}
 & \textbf{3430} & \textbf{2712} & \textbf{2978} & \textbf{9120} \\
\hline
\end{tabular}}
\end{minipage}

\end{table*}

\section{Conclusion and Future Work}  
In this work, we propose a two-stage orientation-aware framework for multi-class brain tumor classification to mitigate domain shift by integrating slice separation with orientation-specific UDA classifiers. While pretrained models fail to generalize across domains, our method achieves a target Macro F1-score of approximately $73\%$, demonstrating effective allignment of MRI distributions. Coronal and sagittal orientations yield the strongest performance, whereas axial slices remain comparatively weaker, likely due to limited discriminative visibility in top-down anatomical views. Despite these improvements, performance is constrained by reduced axial accuracy and class  imbalance. Future work will focus on addressing class imbalance through improved sampling and loss re-weighting strategies to further enhance overall robustness. Additionally, extending the framework to 3D architectures for richer spatial modeling, incorporating advanced adversarial or contrastive adaptation strategies, and validating on larger multi-institutional datasets will be critical for improving clinical generalizability.

%
%
%

\end{document}